\documentclass{article}

\usepackage[final]{neurips_2025}

\usepackage[utf8]{inputenc} 
\usepackage[T1]{fontenc}    
\usepackage{hyperref}       
\usepackage{url}            
\usepackage{booktabs}       
\usepackage{amsfonts}       
\usepackage{nicefrac}       
\usepackage{microtype}      
\usepackage{xcolor}         
\usepackage{graphicx}
\usepackage{bm}
\usepackage{multirow}
\usepackage{amsmath, amssymb}
\usepackage{amsthm}
\usepackage{graphicx}
\usepackage{subcaption}
\newtheorem{definition}{Definition}
\newtheorem{statement}{Statement}
\title{Hybrid-Balance GFlowNet for Solving\\ Vehicle Routing Problems}

\author{
Ni Zhang, 
~Zhiguang Cao\\
School of Computing and Information Systems, Singapore Management University, Singapore\\
\texttt{ni.zhang.2025@phdcs.smu.edu.sg, zgcao@smu.edu.sg}
}

\begin{document}

\maketitle

\begin{abstract}

Existing GFlowNet-based methods for vehicle routing problems (VRPs) typically employ Trajectory Balance (TB) to achieve global optimization but often neglect important aspects of local optimization. While Detailed Balance (DB) addresses local optimization more effectively, it alone falls short in solving VRPs, which inherently require holistic trajectory optimization. To address these limitations, we introduce the Hybrid-Balance GFlowNet (HBG) framework, which uniquely integrates TB and DB in a principled and adaptive manner by aligning their intrinsically complementary strengths. Additionally, we propose a specialized inference strategy for depot-centric scenarios like the Capacitated Vehicle Routing Problem (CVRP), leveraging the depot node's greater flexibility in selecting successors. Despite this specialization, HBG maintains broad applicability, extending effectively to problems without explicit depots, such as the Traveling Salesman Problem (TSP). We evaluate HBG by integrating it into two established GFlowNet-based solvers, i.e., AGFN and GFACS, and demonstrate consistent and significant improvements across both CVRP and TSP, underscoring the enhanced solution quality and generalization afforded by our approach.

\end{abstract}

\section{Introduction}
\label{Introduction}

Vehicle Routing Problems (VRPs) are fundamental to real-world operations, including e-commerce logistics \cite{zhang2021forward,ehrler2021challenges,ranathunga2021solution}, urban delivery \cite{zheng2021urban, choi2019multi, kim2015city}, supply chain management \cite{dondo2011multi,giallanza2020fuzzy,ccil2023integrating}, and ride-sharing systems \cite{guo2022vehicle,lin2012research,tafreshian2020frontiers}. Efficient VRP solutions directly affect cost reduction, service quality, and overall performance in transportation and supply chain networks. Over the past decades, numerous heuristic and meta-heuristic algorithms, such as the Lin-Kernighan-Helsgaun algorithm \cite{helsgaun2000effective}, ant colony optimization (ACO) \cite{bell2004ant}, hybrid genetic search \cite{vidal2022hybrid}, tabu search \cite{barbarosoglu1999tabu}, and simulated annealing \cite{osman1993metastrategy}, have been developed to address the combinatorial complexity of VRPs. However, these approaches often depend on handcrafted rules and problem-specific heuristics, which limit their adaptability and scalability across diverse VRP instances. More recently, reinforcement learning and deep learning methods have emerged as promising alternatives~\cite{pan2023h,bi2022learning,ye2024glop,huang2025rethinking}. Models such as POMO~\cite{kwon2020pomo}, NeuOpt~\cite{ma2024learning}, and DEITSP~\cite{Wang2025} show potential in reducing dependence on handcrafted components. Yet, these methods still struggle to consistently achieve desirable performance, often becoming trapped in local optima due to the limited exploration capacity.

To improve exploration, recent work has explored the use of Generative Flow Network (GFlownet) \cite{bengio2021flow}, which generate diverse and high-quality solutions through a probabilistic, generative process. Unlike traditional learning-based approaches that focus on optimizing a single or a few trajectories, GFlowNet aims to learn a distribution over the solution space, making them well-suited for combinatorial problems like VRPs. However, current GFlowNet-based methods for VRPs such as GFACS \cite{kimant} and AGFN \cite{zhangadversarial}, rely exclusively on global optimization during training. Particularly, they both adopt the Trajectory Balance (TB) objective \cite{malkin2022trajectory}, which effectively aligns with global metrics like minimizing total travel distance. However, this exclusive focus on global optimization can lead to the neglect of important local optimization signals. For instance, these methods may overlook reward dependencies between a current state and its successor, due to the lack of localized training objectives. As a result, they often struggle to capture fine-grained local structures in the solution space, limiting their ability to generate high-quality routes. On the other hand, Detailed Balance (DB) \cite{bengio2021flow} offers a mechanism better suited for local optimization. Nevertheless, using DB alone is equally inadequate, as VRPs fundamentally require a global perspective to achieve optimal solutions. These limitations highlight the need for a broader approach that balances both local and global optimization. We propose that a Hybrid-Balance principle, combining TB and DB in a unified and extensible manner, can significantly enhance GFlowNet-based methods for VRPs.

Guided by this Hybrid-Balance principle, we introduce the Hybrid-Balance GFlowNet (HBG) framework for solving VRPs. First, HBG unifies DB, which promotes local optimization, with TB, which captures global optimization. To fully exploit their complementary strengths, we formulate a VRP-specific version of DB that effectively facilitates local optimization through localized objectives. We also design an adaptive integration mechanism that combines DB and TB in a way that respects their theoretical underpinnings, such as forward and backward transition probability, while leveraging their complementary benefits. Second, motivated by the insight that depot nodes in depot-centric VRPs, such as the Capacitated Vehicle Routing Problem (CVRP), have greater flexibility in selecting successor nodes, we propose a depot-guided inference strategy inspired by the Hybrid-Balance principle. Notably, even in depot-free scenarios like the Traveling Salesman Problem (TSP), our framework remains effective, as the Hybrid-Balance formulation is inherently general. Third, to demonstrate the broad applicability of HBG, we integrate it into two existing GFlowNet-based solvers, i.e., AGFN and GFACS, and observe consistent improvements in routing performance across both CVRP and TSP benchmarks. In summary, our main contributions are outlined as follows:
\begin{itemize}
    \item We propose the Hybrid-Balance GFlowNet (HBG) framework for solving VRPs, which, for the first time, introduces and formalizes the concept of DB within the VRP context. Meanwhile, it unifies the principles of TB and DB through a principled and coherent integration to process both local and global optimizations.

    \item We design a depot-guided inference strategy to efficiently generate and explore high-quality trajectories, specifically tailored for problems involving a designated depot like the CVRP.

    \item We incorporate the HBG framework into existing GFlowNet-based methods for solving VRPs, i.e., AGFN and GFACS, and evaluate it on both synthetic and real-world datasets. The results demonstrate that our method significantly improves the performance of GFlowNet-based solvers for CVRP and TSP.
\end{itemize}

\section{Related Works}
\label{sec:Related Works}
\subsection{Learning-Based Solvers for Vehicle Routing Problems}

Learning-based approaches for VRPs can generally be divided into two categories: construction-based and improvement-based methods. Construction-based solvers generate complete solution trajectories in an end-to-end manner. A seminal example is the Attention Model (AM) \cite{koolattention}, which first applied a Transformer architecture to solve VRPs. Building on AM, Policy Optimization with Multiple Optima (POMO) extends this approach by leveraging multiple optimal policies during training and inference to improve both solution quality and robustness. This line of work has since inspired a series of end-to-end construction-based methods \cite{sun2023difusco, kwon2021matrix, fang2024invit, Wang2025, gao2024towards} that further improve performance. Improvement-based methods, on the other hand, enhance initial solutions through iterative refinements. These methods often integrate neural networks into classical heuristic frameworks. Notable examples include NeuroLKH \cite{xin2021neurolkh}, DeepACO \cite{ye2023deepaco}, and NeuOpt~\cite{ma2024learning}, which demonstrate strong performance through learning-augmented optimization strategies. To showcase the generality of our proposed framework, we apply it to enhance both a construction-based solver (AGFN) and an improvement-based solver (GFACS).
\subsection{GFlowNet for Combinatorial Optimization Problems}

GFlowNet has been applied across a wide range of structured generation and decision-making tasks. In molecular and drug discovery \cite{zhu2023sample, nica2022evaluating, koziarski2024rgfn, shen2023tacogfn,gopalan2025generative,seo2024generative}, they are used to sample diverse, high-reward molecules from complex solution spaces. In causal structure learning \cite{li2022gflowcausal, da2023human}, GFlowNet facilitates exploration over multiple plausible directed acyclic graphs (DAGs), while in Bayesian inference \cite{deleu2022bayesian,silva2024streaming,nishikawa2022bayesian}, they serve as alternative samplers for discrete posteriors. Additional applications include symbolic reasoning \cite{thomas2023neuro, li2023gfn}, robotics planning \cite{licflownets,nagiredla2024robonet}, and solving maximum independent set (MIS) \cite{zhang2023let}, where modeling solution diversity is essential. Recently, GFlowNet has also been applied to VRPs \cite{zhangadversarial,kimant}, including TSP and CVRP. In this context, learning a distribution over feasible routes offers a flexible and effective alternative to deterministic solvers. Two representative models are AGFN and GFACS. AGFN incorporates adversarial training to improve trajectory construction in an end-to-end fashion, making it the first to apply GFlowNet to VRPs directly. In contrast, GFACS integrates GFlowNet with ant colony optimization, marking the first attempt to augment heuristic search with GFlowNet-based learning. In this paper, we further enhance both AGFN and GFACS for solving VRPs using our proposed Hybrid-Balance GFlowNet framework.
\section{Hybrid-Balance GFlowNet}
\label{Hybrid-Balance GFlowNet}
\begin{figure*}[!htb]
    \centering
    \includegraphics[width=\textwidth, trim=0cm 6cm 5cm 0cm, clip]{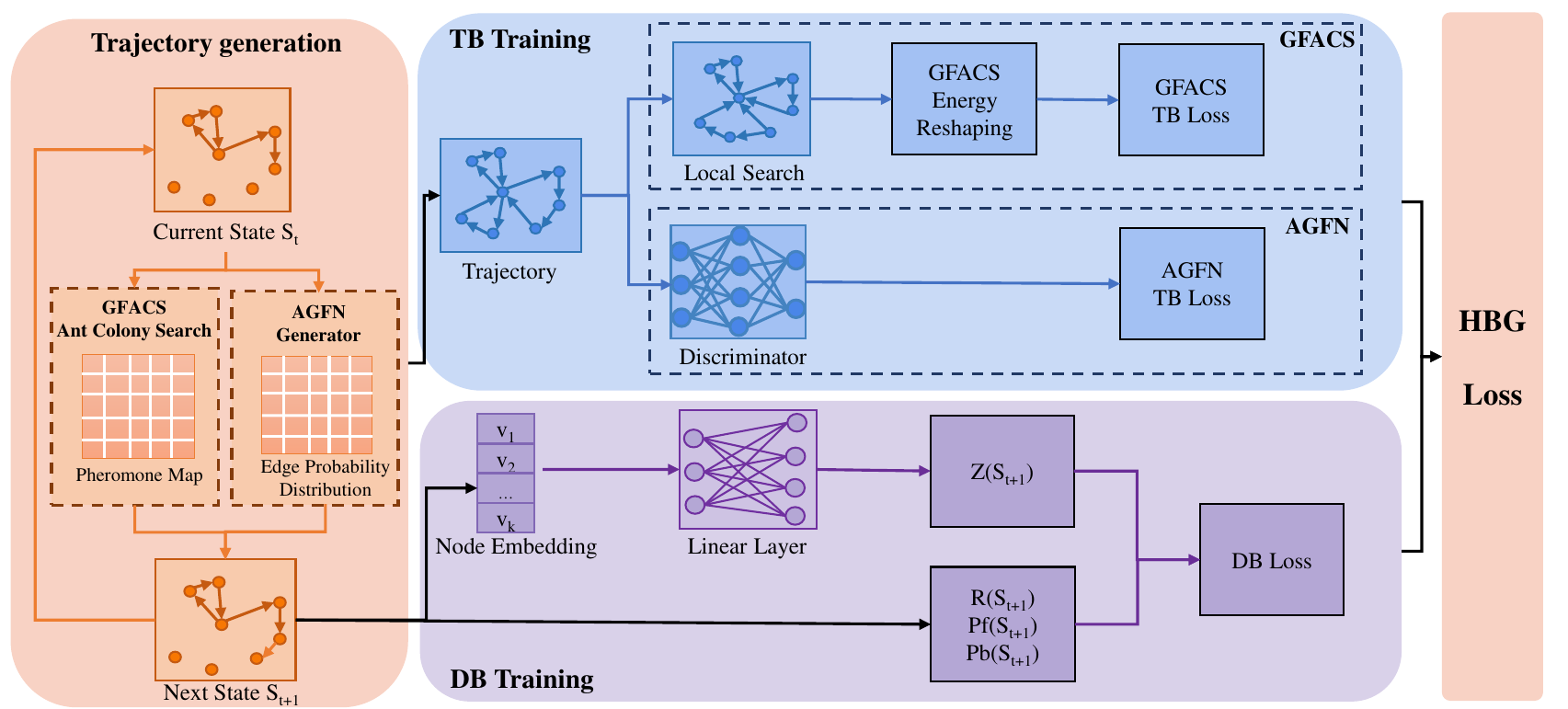}
    \caption{The Overall Framework of Our Hybrid-Balance GFlowNet for Solving VRPs.}
    \label{fig:Framework}
\end{figure*}

As illustrated in Fig.\ref{fig:Framework}, the Hybrid-Balance GFlowNet (HBG) begins with trajectory generation using either AGFN or GFACS, during which state transition information is recorded at each step. The generated next state is then treated as the current state in the following step, and this process repeats until a complete trajectory is constructed. The blue region in Fig.\ref{fig:Framework} corresponds to the original components from AGFN and GFACS, responsible for the processing of the complete trajectory and the computation of the TB loss, which captures global optimization signals. However, relying solely on the TB loss can cause the model to overlook important relationships between individual states. To bridge this gap, our proposed HBG introduces additional components, highlighted in purple, where the DB loss is computed for each individual state transition to enhance local optimization. The final training objective is a combination of two loss, which together guide the optimization of the model.

To better illustrate the motivation for introducing DB, consider a long vehicle routing trajectory that is incrementally constructed as A → B → C → D → E → ... → U → V → W → X → Y → Z. Assume this complete route yields a high total cost (bad performance), primarily due to suboptimal decisions made in the early stages, such as traversing a high-cost edge from node C to node D. In contrast, the latter portion of the tour (e.g., from node U to Z) may follow a more cost-effective and well-structured pattern. Under Trajectory Balance (TB), the final reward is determined by the overall trajectory cost, and is proportionally assigned to all transitions. Consequently, even high-quality local transitions, such as W → X → Y, may receive weak or misleading training signals simply because they are embedded in a globally suboptimal trajectory. This would hinder the model's ability to learn and reinforce desirable local patterns. On the other hand, Detailed Balance (DB) operates at a step-wise granularity, evaluating the expected outcomes of individual transitions. For instance, at node W, DB can assess whether transitioning to X leads to better outcomes compared to alternative choices like Z, regardless of earlier suboptimal steps. This localized and reward-sensitive feedback enables the model to more accurately learn local quality from global performance, and promotes stronger learning signals for valuable decisions even within imperfect trajectories.

This example illustrates a core limitation of Trajectory Balance (TB) in long-horizon combinatorial tasks like VRP: when the overall trajectory is suboptimal, TB lacks the ability to identify and preserve well-structured local segments within it. As a result, valuable local patterns may be overlooked or penalized. By incorporating Detailed Balance (DB) into the training objective, we address this limitation by providing fine-grained, step-level singal that helps isolate and reinforce high-quality local decisions, even when the global trajectory does not show good performance.

\subsection{Modeling Basics}
\textbf{Problem Definition. }
For a CVRP instance, $\mathcal{G}$ denotes the input graph, which includes the coordinates and demands of customers, as well as the depot location. Formally, the instance is represented as a complete graph $\mathcal{G} = (\mathcal{V}, \mathcal{U})$, where $\mathcal{V} = \{v_0, v_1, \dots, v_n\}$ denotes the set of nodes, with $v_0$ as the depot and the remaining nodes representing customers, and $\mathcal{U}$ is the set of edges. Each customer node $v_i$ ($i \geq 1$) is associated with a demand $d_i$ and a location in Euclidean space. Each edge $(v_i, v_j) \in \mathcal{U}$ has an associated cost $c_{ij}$, typically defined as the Euclidean distance between $v_i$ and $v_j$. The goal of CVRP is to determine a set of vehicle routes that start and end at the depot, such that each customer is visited exactly once, the total demand on each route does not exceed the vehicle's capacity, and the total routing cost is minimized.

\noindent\textbf{State $s$:} In a trajectory set $\mathcal{T} = \{\tau_1, \tau_2, \ldots, \tau_h\}$, the state $s^i$ denotes the sequence of nodes visited in trajectory $\tau_i$. At decision step $t$ in $\tau_i$, the state is defined as $s_t^i = \{x_0^i, x_1^i, x_2^i, \dots, x_t^i\}$, where $x_t^i$ is the most recently visited node, and $x_0^i$ represents the depot which serves as both the starting and ending point of the route.

\noindent\textbf{Action $a$:} An action $a_{t}^i$ transitions the system from state $s_t^i$ to $s_{t+1}^i$. Given $s_t^i = \{x_0^i, x_1^i, \dots, x_t^i\}$, the action selects the next node $x_{t+1}^i$ from the set of unvisited nodes, adhering to feasibility constraints such as vehicle capacity. Once all customers are visited, the route terminates in a final state, forming a complete trajectory $\tau_i = \{x_0^i, x_1^i, \dots, x_m^i\}$.

\noindent\textbf{Reward $R$:} The reward $R(\tau_i)$ is determined by the quality of the generated trajectory $\tau_i$. We define two types of rewards: $R(\tau_i)$ and $R(s_t^i)$. The former, $R(\tau_i)$, evaluates the entire trajectory, while the latter, $R(s_t^i)$, reflects local reward signals at individual state transitions. These are defined as: $R(\tau_i) = \sum_{k=0}^{m-1} d(x_k^i, x_{k+1}^i)$, $R(s_t^i) = d(x_{t-1}^i, x_t^i)$, where $d(x_k^i, x_{k+1}^i)$ denotes the Euclidean distance between consecutive nodes.

\textbf{Graph Neural Network (GNN).} 
We integrate a GNN module \cite{wu2020comprehensive} into the GFlowNet framework to more effectively capture the complex relational structures inherent in VRP instances. The detailed architecture and formulation are provided in Appendix~\ref{sec:GNN}. Following the designs of AGFN and GFACS, we sparsify the fully connected graph into a $k$-nearest-neighbor graph $\mathcal{G}^{}$ to improve scalability and reduce computational cost. The graph $\mathcal{G}^{}$ is embedded into a high-dimensional feature space, encoding node coordinates and edge distances as node and edge features, respectively. The GNN, parameterized by $\bm{\theta}$, processes these features through multiple layers to produce rich representations. The resulting edge embeddings are passed through a multi-layer perceptron (MLP) to generate edge probability distribution $\eta(\mathcal{G}^{*}, \bm{\theta})$ for decision making by GFlowNet, while the node embeddings $\mathcal{Q} = \{q_1, q_2, \ldots, q_b\}$ are retained for computing state flows.

\subsection{Hybrid-Balance GFlowNet}
\subsubsection{Global Optimization via TB}


In VRPs, the objective is to determine the shortest route while satisfying various operational constraints, which necessitates evaluating solutions from a global perspective. Both AGFN and GFACS adopt the Trajectory Balance (TB) objective to address this requirement, as it enables the GFlowNet to be trained over entire trajectories, naturally aligning with global optimization goals.

As illustrated in Fig.~\ref{fig:Framework}, AGFN generates an edge probability distribution $\eta(\mathcal{G}^{*}, \bm{\theta}_\text{generator})$ using GFlowNet, which is then used to sample the next node in the route. A discriminator, trained with false labels from GFlowNet-generated trajectories and true labels from near-optimal trajectories, evaluates the quality of sampled trajectory set $\mathcal{T} = \{\tau_1, \tau_2, \ldots, \tau_h\}$. It assigns a quality score to each trajectory, which is then combined with the raw trajectory length $R(\tau)$ to compute the final AGFN reward $\widetilde{R}(\tau)$. These rewards $\widetilde{R}(\tau)$, along with the source flow $Z(\bm{\theta}_\text{generator})$, forward probability $P_F(\tau; \bm{\theta}_\text{generator})$, and backward probability $P_B(\tau)$ obtained from the GFlowNet, are used to compute the AGFN TB loss $\ell^{\text{AG}}_{\text{TB}}$, defined as:
\begin{equation}
    \label{eq:AGFN loss function}
    \ell^{\text{AG}}_{\text{TB}}(\mathcal{T}; \bm{\theta}_\text{generator}) = \frac{1}{h} \sum_{k=1}^{h}\left( \log \frac{Z(\bm{\theta}_\text{generator})*P_{F}({\tau}_{k}; \bm{\theta}_\text{generator})}{\widetilde{R}({\tau}_{k})* P_{B}({\tau}_{k})} \right)^2.
\end{equation}


For GFACS, the GFlowNet is used to generate a heuristic matrix $\eta(\mathcal{G}^{*}, \bm{\theta})$, which is subsequently transformed into a pheromone map to guide the ant colony optimization (ACO) in trajectory construction. Once the trajectories are generated, a local search is applied for refinement, followed by an energy reshaping step. The TB loss for GFACS, denoted $\ell^{\text{GF}}_{\text{TB}}$, is then computed in a similar form to AGFN, as both approaches adopt the TB loss formulation to optimize their models.

\subsubsection{Local-Global Optimization through Hybrid-Balance}
While global optimization is essential for solving VRPs, local optimization is also important as it helps the model to capture fine-grained patterns, such as transitions between neighboring nodes. However, local information alone is insufficient for modeling global objective and constraints like total cost and capacity. To address this, we propose to unify both global and local objectives within a Hybrid-Balance GFlowNet framework. Specifically, we integrate the DB mechanism into the original TB framework of the GFACS and AGFN models to further enhance the modeling of local transitions, particularly the relationship between the current state $s_t^i$ and the next state $s_{t+1}^i$.


As shown in Fig.~\ref{fig:Framework}, the model records relevant information at each step, including the current state's reward $R(s_t^i)$ and flow $F(s_t^i;\bm{\theta})$, the next state's reward $R(s_{t+1}^i)$ and flow $F(s_{t+1}^i;\bm{\theta})$, as well as the forward and backward transition probability $P_f(s_{t+1}^i|s_t^i; \bm{\theta})$ and $P_b(s_t^i|s_{t+1}^i)$. Once a trajectory is completed, we apply a forward-looking technique \cite{pan2023better} to compute the DB loss $\ell_{\text{DB}}$ between two successive states as:

\begin{equation}
\label{eq:DB function}
\ell_{\text{DB}}(s_t^i, s_{t+1}^i; \bm{\theta}) = \left( \log \frac{P_f(s_{t+1}^i|s_t^i; \bm{\theta}) \cdot F(s_t^i; \bm{\theta}) \cdot \exp(\tilde{\mathcal{E}}(s_{t+1}^i))}{P_b(s_t^i|s_{t+1}^i) \cdot F(s_{t+1}^i; \bm{\theta}) \cdot \exp(\tilde{\mathcal{E}}(s_t^i))} \right)^2.
\end{equation}

Here, $P_f(s_{t+1}^i|s_t^i; \bm{\theta})$ denotes the forward transition probability derived from the edge probability distribution $\eta(\mathcal{G}^{*}, \bm{\theta})$ in AGFN or the pheromone map in GFACS. The relationship between the trajectory-level forward probability $P_{F}(\tau_i; \bm{\theta})$ used in TB loss and the step-wise forward probability $P_f(s_{t+1}^i|s_t^i; \bm{\theta})$ used in DB loss is given by:

\begin{equation}
    \label{eq:forward}
    P_{F}({\tau_i}; \bm{\theta})=\prod_{t=1}^{m} P_{f}(s_t^i|s_{t-1}^i; \bm{\theta}).
\end{equation}

To ensure consistency with the trajectory-level backward probability $P_B(\tau_i)$ used in TB loss, we design the step-wise backward probability $P_b(s_t^i|s_{t+1}^i)$ to reflect the structure of sub-trajectories within $\tau_i$. Specifically, we assume that each complete trajectory $\tau_i$ consists of $a$ multi-node sub-trajectories and $j$ single-node sub-trajectories, and parameter $P_b$ is accordingly determined by the varied transition structures.

\begin{definition}[Trajectory Composition and Ordering Count]
We define $\mathcal{A}_a$ as the set of $a$ multi-node trajectories, and $\mathcal{J}_j$ as the set of $j$ single-node trajectories. Together, these sequences are combined to form a complete trajectory $\tau_i$. Let $B(\mathcal{A}_a, \mathcal{J}_j)$ denote the number of distinct orderings of sub-trajectories in $\mathcal{A}_a$ and $\mathcal{J}_j$ that result in the same complete trajectory $\tau_i$.
\end{definition}
We next present the following statement, which describes the recurrence relation for $B(\mathcal{A}_a, \mathcal{J}_j)$.
\begin{statement}[Trajectory Orders' Count Recurrence]
The number of distinct trajectories composed of $a$ multi-node trajectories and $j$ single-node trajectories arranged in different orders, denoted by $B(\mathcal{A}_a, \mathcal{J}_j)$, satisfies the following recurrence relation:
\begin{equation}
\label{eq. Trajectory Count Recurrence}
B(\mathcal{A}_a, \mathcal{J}_j) = 2a \cdot B(\mathcal{A}_{a-1}, \mathcal{J}_j) + j \cdot B(\mathcal{A}_a, \mathcal{J}_{j-1}).
\end{equation}
\end{statement}
This recurrence arises from the backward destruction of CVRP trajectories, where we consider how $B(\mathcal{A}_a, \mathcal{J}_j)$ reached its predecessors. Suppose the current state corresponds to $B(\mathcal{A}_a, \mathcal{J}_j)$, where there are $a$ remaining multi-node trajectories and $j$ remaining single-node trajectories to be disconnected. There are two possible types of backward transitions from this state to reach its predecessor $B(\mathcal{A}_{a-1}, \mathcal{J}_j)$ or $B(\mathcal{A}_a, \mathcal{J}_{j-1})$:

\textbf{(1) Multi-node trajectory:} If a multi-node trajectory is selected for backward destruction from the depot, either of its two nodes can serve as the immediate predecessor to the depot. Therefore, each of the $a$ multi-node trajectories contributes two valid backward transitions, resulting in a total contribution of $2a \cdot B(\mathcal{A}_{a-1}, \mathcal{J}_j)$, where the recursion proceeds with $a-1$ remaining multi-node trajectories and $j$ unchanged single-node trajectories.

\textbf{(2) Single-node trajectory:} If a single-node trajectory is chosen, it contains only one node, which uniquely determines the depot’s predecessor. Thus, each of the $j$ single-node trajectories contributes one backward transition, resulting in $j \cdot B(\mathcal{A}_a, \mathcal{J}_{j-1})$, where the recursion continues with $a$ multi-node trajectories and $j-1$ single-node trajectories.

We combine both types of transitions to derive the recurrence relation as presented in Eq.~\ref{eq. Trajectory Count Recurrence}. Subsequently, we deduce the closed-form expression of $B(\mathcal{A}_a, \mathcal{J}_j)$ from Eq.~\ref{eq. Trajectory Count Recurrence}. The proof is provided in Appendix Sec.~\ref{sec:Proof Process of Hybrid-Balance}, and the corresponding formulation is presented below:
\begin{equation}
B(\mathcal{A}_a, \mathcal{J}_j) = (a + j)! \cdot 2^a, \quad \text{for } a, j \geq 0.
\end{equation}
Physically, the term $(a + j)!$ accounts for all possible orderings of the $a + j$ sub-trajectories, where each of them is treated as an atomic step in the destruction process. Each multi-node trajectory has 2 possible directions for destruction, contributing an additional $2^a$ multiplicative factor. In contrast, single-node trajectories allow only one valid direction. Therefore, the total number of reward-equivalent permutations is the product of these two factors.

\begin{statement}[TB Backward Probability]
We denote $P_B(\tau_i)$ as the backward policy probability of a complete trajectory $\tau_i$ in the GFlowNet framework under TB, formulated as:
\begin{equation}
    P_B(\tau_i) = \frac{1}{(a + j)! \cdot 2^a},
\end{equation}
\end{statement}
where the denominator reflects the total number of distinguishable trajectory permutations given $a$ multi-node and $j$ single-node trajectories to achieve complete trajectory $\tau_i$.

\begin{statement}[DB Backward Probability]
We denote $P_b(s_t^i \mid s_{t+1}^i)$ as the probability of a single backward transition from state $s_{t+1}^i$ to its predecessor $s_t^i$, and under the DB formulation, the backward probability is defined conditionally:
\begin{equation}
    P_b(s_t^i|s_{t+1}^i) = 
    \begin{cases}
    \frac{1}{2a + j} & \text{if the current node is the depot}, \\
    1 & \text{otherwise}.
    \end{cases}
\end{equation}
\end{statement}
This probability formulation originates from Eq.~\ref{eq. Trajectory Count Recurrence}, which defines the total number of sub-trajectory backward destruction orderings. Physically, each multi-node trajectory offers two possible predecessor nodes for backward disconnection from the depot, thereby contributing the $2a$ term, while each single-node trajectory provides one such option, contributing the $j$ term. The resulting probability $\frac{1}{2a + j}$ reflects a uniform selection over all valid backward transitions at the current decision step. In contrast, for all other nodes in the trajectory, only a single predecessor is feasible, and thus the backward transition becomes fully deterministic with probability $1$.

Meanwhile, $F(s_{t}^i;\bm{\theta})$ in Eq. \ref{eq:DB function} represents the flow of current state, and is derived from the node embedding $q$ at state $s_t^i$, which is calculated as follow:
\begin{equation}
F(s_{t}^i;\bm{\theta}) = \frac{1}{t} \sum_{x_k \in s_t^i} (W_2 \cdot \text{ReLU}(W_1 \cdot q_k + b_1) + b_2),
\end{equation}
where $W_1$, $W_2$, $b_1$ and $b_2$ are learnable parameters and ReLU \cite{chen2020dynamic} is the activation function. To handle the local objective associated with state transitions, we define the reward of the predecessor state $s_t^i$ as zero. Consequently, the energy term $\tilde{\mathcal{E}}(s_t^i)$ in Eq.~\ref{eq:DB function} is also set to zero. The successor state $s_{t+1}^i$, in contrast, receives a non-zero transition reward. Accordingly, the energy term $\tilde{\mathcal{E}}(s_{t+1}^i)$ represents the local reward signal, and its negative is defined as follow:
\begin{equation}
\label{eq:energy function}
    \tilde{\mathcal{E}}(s_t^i)= R({s}_{t}^i)-\frac{1}{h}\sum_{k=1}^{h}R({s}_{t}^k).
\end{equation}
As training progresses, the quality of each trajectory steadily improves, resulting in smaller values of $R(s_t^i)$ as the generated routes become shorter. In Eq. \ref{eq:energy function}, we compute the energy $\tilde{\mathcal{E}}(s_t^i)$ for state $s_t^i$ by subtracting the average reward of other trajectories at the same decision step. This formulation effectively captures the relative advantage of a given state compared to its peers, encouraging the model to assign higher energy to better-performing states. As the variance across rewards decreases during training, the energy values naturally increase.

Then, the DB loss of the completed trajectory $\tau_i= \{x_0^i, x_1^i, x_2^i, \dots, x_m^i\}$ can be calculated as:
\begin{equation}
    \ell_{\text{DB}}(\tau_i; \bm{\theta}) =\sum_{t=0}^{m-1}\ell_{\text{DB}}(s_t^i, s_{t+1}^i; \bm{\theta}), 
\end{equation}
where $\ell_{\text{DB}}(s_t^i, s_{t+1}^i; \bm{\theta})$ is derived from Eq. \ref{eq:DB function}. The overall loss for the Hybrid-Balance GFlowNet, denoted by $\ell_{\text{HB}}(\mathcal{T}; \bm{\theta})$, is computed by aggregating both the TB loss $\ell_{\text{TB}}(\tau; \bm{\theta})$ and the DB loss $\ell_{\text{DB}}(\tau; \bm{\theta})$ over all trajectories:
\begin{equation}
    \ell_{\text{HB}}(\mathcal{T}; \bm{\theta})=\sum_{i=1}^{h}\ell_{\text{HB}}(\tau_i; \bm{\theta}) =\sum_{i=1}^{h}(\ell_{\text{TB}}(\tau_i; \bm{\theta})+\ell_{\text{DB}}(\tau_i; \bm{\theta})).
\end{equation}
This unified objective enables the model to simultaneously capture global trajectory-level structure and fine-grained local transitions, leading to more effective and robust optimization in VRPs.

\subsubsection{Depot-Guided Inference}
\begin{figure*}[!htb]
    \centering
    \includegraphics[width=\textwidth, trim=6cm 9.5cm 1.6cm 2cm, clip]{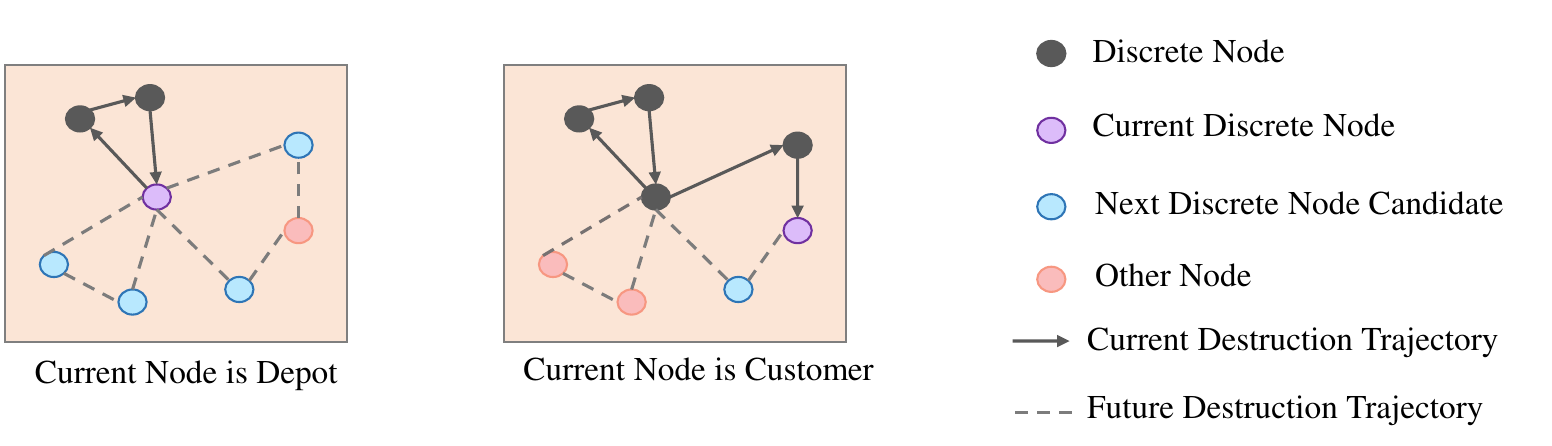}
    \caption{Illustration for Depot-Guide Inference.}
    \label{fig:Depot-Guide Inference}
\end{figure*}
The design of Hybrid-Balance GFlowNet's backward policy reveals a key insight: as illustrated in Fig.~\ref{fig:Depot-Guide Inference}, only the depot node retains flexibility in choosing among multiple predecessor candidates during trajectory destruction. This flexibility stems from the construction of sub-trajectories, each of which begins and ends at the depot. In contrast, for all customer nodes, the backward transition path is uniquely defined by the trajectory structure, i.e., once a customer node is reached, its predecessor is deterministically identified. This determinism also holds during forward trajectory construction. To leverage this structural characteristic, we propose a depot-guided inference mechanism defined as:
\begin{equation}
    x_{t+1}=\begin{cases}
    x & \text{if the current node $x_t$ is depot}, \\
    x^* & \text{if the current node $x_t$ is customer},
    \end{cases}
\end{equation}

where $x \sim P_f(s_{t+1} \mid s_t; \bm{\theta})$ denotes sampling from the forward transition probability, and $x^* = \arg\max P_f(s_{t+1} \mid s_t; \bm{\theta})$ corresponds to greedy selection. Here, $P_f(s_{t+1} \mid s_t; \bm{\theta})$ is derived from the edge probability distribution $\eta(\mathcal{G}^{*}, \bm{\theta})$ in AGFN or the pheromone map in GFACS. Under this strategy, exploration through sampling is applied only at the depot, while customer nodes follow a deterministic, greedy policy.


It is important to note that depot-guided inference is specifically designed for problems featuring a designated depot node, such as the CVRP. For problems lacking a depot or node-role differentiation, such as the TSP, we retain their original inference procedures, including hybrid decoding strategy~\cite{zhangadversarial} for AGFN and the ant clony search \cite{kimant} for GFACS.


\section{Experiment}
We conduct experiments to validate the effectiveness of the Hybrid-Balance GFlowNet (HBG) in enhancing two representative GFlowNet-based solvers, i.e., AGFN and GFACS, on CVRP. We first present comparison results, followed by ablation studies to analyze the contribution of individual components. Lastly, we extend the evaluation to other vehicle routing problem.

\textbf{Dataset:} We adopt synthetic CVRP datasets following standard settings used in prior work \citep{kimant,kwon2020pomo,zhangadversarial,xin2021neurolkh}. Each instance features a single depot and multiple customers served by a vehicle with fixed capacity $C$. The depot and customer coordinates are sampled uniformly from the unit square $[0,1]^2$, and customer demands follow a uniform distribution $U[a, b]$ with $a = 1$ and $b = 9$. The vehicle capacity is fixed at $C = 50$ across all problem sizes: 100, 200, 500, and 1,000 nodes. For testing, we generate 128 synthetic instances for each of the 200-, 500-, and 1,000-node settings, aligned with evaluation rules established in AGFN and GFACS. The code is available at \url{https://github.com/ZHANG-NI/HBG}

\textbf{Hyperparameters:} We adopt the same model configurations and training settings as AGFN and GFACS, including network architecture, batch size, learning rate, optimizer, and other hyperparameters. Training is conducted using sampling-based decoding with $\mathcal{N} = 20$ routes per instance. During inference, AGFN uses depot-guided inference, and GFACS applies an ant colony search with depot-guided node selection. All models are trained on 100-node instances. The experiments are conducted on a server equipped with an NVIDIA A100 GPU and an Intel Xeon 6342 CPU.
\subsection{Performance on Synthetic CVRP Instances}
\begin{table*}[t]
\centering
\caption{Comparison on CVRP datasets of different sizes: Objective (Obj.) values and inference times (in seconds) are shown, and Gap(\%) is computed with respect to LKH.}
\resizebox{\textwidth}{!}{
\begin{tabular}{l|ccc|ccc|ccc}
\toprule
\multirow{2}{*}{\textbf{Method}} & \multicolumn{3}{c|}{\textbf{200}} & \multicolumn{3}{c|}{\textbf{500}} & \multicolumn{3}{c}{\textbf{1000}} \\
& \textbf{Obj. $\downarrow$} & \textbf{Time(s) $\downarrow$} & \textbf{Gap(\%) $\downarrow$} & \textbf{Obj. $\downarrow$} & \textbf{Time(s) $\downarrow$} & \textbf{Gap(\%) $\downarrow$} & \textbf{Obj. $\downarrow$} & \textbf{Time(s) $\downarrow$} & \textbf{Gap(\%) $\downarrow$} \\
\midrule
LKH & 28.04 & 59.81 & - & 63.32 & 233.72 & - & 120.53 & 433.90 & - \\
ACO &71.46 &3.36 & 154.85 &189.79 &11.14 & 199.73 &371.30 &24.50 & 208.06\\
\midrule
POMO (*8) & 29.22 & 0.29 & 4.21 & 79.86 & 0.84 & 26.12 & 192.18 & 3.06 & 59.45 \\
POMO & 29.45 & 0.23 & 5.03 & 82.92 & 0.59 & 30.95 & 231.88 & 1.48 & 92.38 \\
GANCO & 29.01 & 0.46 & 3.57 & 71.30 & 148.91 & 12.60 & 145.84 & 4.02 & 21.00 \\
NeuOpt & 38.42 &17.19 & 37.02 & 186.17 &38.05 & 194.01 & - &- & -\\
\midrule
AGFN & 31.26 & 0.14 & 11.48 & 71.05 & 0.40 & 12.21 & 133.97 & 0.65 & 11.15 \\
HBG-AGFN & \textbf{30.83} & 0.15 & \textbf{9.95} & \textbf{69.93} & 0.42 & \textbf{10.44} & \textbf{131.78} & 0.65 & \textbf{9.34} \\
GFACS & 34.52 & 4.65 & 23.11 & 78.41 & 12.76 & 23.83 & 149.24 & 26.32 & 23.82 \\
HBG-GFACS & \textbf{32.66} & 4.67 & \textbf{16.48} & \textbf{71.89} & 12.77 & \textbf{13.53} & \textbf{133.32} & 26.33 & \textbf{10.61} \\
GFACS (local search) & 28.63 & 12.18 & 2.10 & 65.24 & 34.19 & 3.03 & 124.15 & 80.52 & 3.00 \\
HBG-GFACS (local search) & \textbf{28.59} & 12.20 & \textbf{1.96} & \textbf{65.10} & 34.21 & \textbf{2.81} & \textbf{123.85} & 80.53 & \textbf{2.75} \\
\bottomrule
\end{tabular}
}
\vspace{-4mm}
\label{tab:cvrp_results}
\end{table*}

We compare HBG-enhanced models, i.e., HBG-AGFN and HBG-GFACS, with their original TB-based counterparts, AGFN \cite{zhangadversarial} and GFACS \cite{kimant}. AGFN constructs routes in an end-to-end manner, while GFACS searches for solutions by combining GFlowNet with ant colony optimization. We also include classical heuristics (LKH \cite{helsgaun2000effective}, ACO \cite{bell2004ant}) and learning-based baselines (POMO \cite{kwon2020pomo}, GANCO \cite{xin2022generative}, NeuOpt \cite{ma2024learning}) for comparison. All methods are trained on 100-node instances and evaluated on CVRP200, CVRP500, and CVRP1000 datasets, following AGFN and GFACS evaluation protocols. Additional experiments on the public benchmark CVRPLib are reported in Appendix~\ref{sec:Test on Real world dataset}.


Table~\ref{tab:cvrp_results} shows that HBG consistently improves performance across all problem sizes for AGFN, GFACS, and GFACS with local search. The performance gains are significant, with gap reductions of up to 16.23\%, 55.46\%, and 8.33\%, respectively. Improvements become more pronounced as instance size increases, indicating strong scalability. Inference incurs only minor overhead (0.01–0.04 seconds) due to temporary loading of flow parameters, which does not impact overall runtime or scalability. Compared to other heuristic and learning-based methods, HBG-AGFN and HBG-GFACS achieve competitive or superior solution quality across all scales. On CVRP200, both methods outperform ACO and NeuOpt. On CVRP500 and CVRP1000, they continue to generalize effectively, outperforming ACO, POMO, GANCO, and NeuOpt. These results highlight the robustness, efficiency, and strong generalization capabilities of the proposed HBG framework.

\begin{table*}[t]
\centering
\caption{Ablation Study on AGFN and GFACS: Gap(\%) is computed with respect to LKH.}
\begin{minipage}{0.49\textwidth}
\centering
\subcaption{AGFN}
\resizebox{\textwidth}{!}{
\begin{tabular}{l|cc|cc|cc}
\toprule
\multirow{2}{*}{\textbf{Method}} & \multicolumn{2}{c|}{\textbf{200}} & \multicolumn{2}{c|}{\textbf{500}} & \multicolumn{2}{c}{\textbf{1000}} \\
& \textbf{Obj. $\downarrow$} & \textbf{Gap(\%) $\downarrow$} & \textbf{Obj. $\downarrow$}  & \textbf{Gap(\%) $\downarrow$} & \textbf{Obj. $\downarrow$}  & \textbf{Gap(\%) $\downarrow$} \\
\midrule
LKH & 28.04 & - & 63.32 & - & 120.53  & - \\
\midrule
AGFN & 31.26 & 11.48 & 71.05 & 12.21 & 133.97 & 11.15\\
~~~~+ HB & 31.08 & 10.84 & 69.99 & 10.53 & 131.94 & 9.47\\
~~~~~~~~+ Depot-Guide Inference & 30.83 & 9.95 & 69.93 & 10.44 & 131.78 & 9.34\\
\bottomrule
\end{tabular}
}
\label{tab:agfn_ablation}
\end{minipage}
\hfill
\begin{minipage}{0.49\textwidth}
\centering
\subcaption{GFACS}
\resizebox{\textwidth}{!}{
\begin{tabular}{l|cc|cc|cc}
\toprule
\multirow{2}{*}{\textbf{Method}} & \multicolumn{2}{c|}{\textbf{200}} & \multicolumn{2}{c|}{\textbf{500}} & \multicolumn{2}{c}{\textbf{1000}} \\
& \textbf{Obj. $\downarrow$} & \textbf{Gap(\%) $\downarrow$} & \textbf{Obj. $\downarrow$}  & \textbf{Gap(\%) $\downarrow$} & \textbf{Obj. $\downarrow$}  & \textbf{Gap(\%) $\downarrow$} \\
\midrule
LKH & 28.04 & - & 63.32 & - & 120.53  & - \\
\midrule
GFACS & 34.52 & 23.11 & 78.41 & 23.83 & 149.24 & 23.82\\
~~~~+ HB & 34.01 & 21.29 & 76.67 & 21.08 & 144.48 & 19.87\\
~~~~~~~~+ Depot-Guide Inference & 32.66 & 16.48 & 71.89 & 13.53 & 133.32 & 10.61\\
\bottomrule
\end{tabular}
}
\end{minipage}
\label{tab:HBG_ablation}
\end{table*}

\subsection{Ablation Study}
\paragraph{Comparison of Component Contributions.} We evaluate the contribution of each component in the HBG framework for both AGFN and GFACS. First, we incorporate the Hybrid-Balance (HB) module into the original models. Then, we add the depot-guided inference mechanism on top of the HB-enhanced variants. As shown in Table~\ref{tab:HBG_ablation}, each component contributes significantly to performance. Incorporating the HB module alone reduces the optimality gap by up to 15.07\% in AGFN and 16.58\% in GFACS. Adding depot-guided inference provides further gains, especially for larger instances. These results confirm that the HB module offers consistent improvements and depot-guided inference delivers additional benefits in depot-centric tasks.

\begin{table*}[t]
\centering
\caption{Comparison of DB, TB, and HB: Gap(\%) is computed with respect to LKH.}
\begin{minipage}{0.465\textwidth}
\centering
\subcaption{AGFN}
\resizebox{\textwidth}{!}{
\begin{tabular}{l|cc|cc|cc}
\toprule
\multirow{2}{*}{\textbf{Method}} & \multicolumn{2}{c|}{\textbf{200}} & \multicolumn{2}{c|}{\textbf{500}} & \multicolumn{2}{c}{\textbf{1000}} \\
& \textbf{Obj.} & \textbf{Gap(\%)} & \textbf{Obj.}  & \textbf{Gap(\%)} & \textbf{Obj.}  & \textbf{Gap(\%)} \\
\midrule
LKH & 28.04 & -- & 63.32 & -- & 120.53 & -- \\
\midrule
DB & 34.41 & 22.72 & 76.78 & 21.26 & 143.25 & 18.85 \\
TB & 31.26 & 11.48 & 71.05 & 12.21 & 133.97 & 11.15 \\
HB & \textbf{31.08} & \textbf{10.84} & \textbf{69.99} & \textbf{10.53} & \textbf{131.94} & \textbf{9.47} \\
\bottomrule
\end{tabular}
}
\label{tab:agfn_ablation}
\end{minipage}
\hfill
\begin{minipage}{0.515\textwidth}
\centering
\subcaption{GFACS}
\resizebox{\textwidth}{!}{
\begin{tabular}{l|cc|cc|cc}
\toprule
\multirow{2}{*}{\textbf{Method}} & \multicolumn{2}{c|}{\textbf{200}} & \multicolumn{2}{c|}{\textbf{500}} & \multicolumn{2}{c}{\textbf{1000}} \\
& \textbf{Obj. $\downarrow$} & \textbf{Gap(\%) $\downarrow$} & \textbf{Obj. $\downarrow$}  & \textbf{Gap(\%) $\downarrow$} & \textbf{Obj. $\downarrow$}  & \textbf{Gap(\%) $\downarrow$} \\
\midrule
LKH  & 28.04 & -- & 63.32 & -- & 120.53 & -- \\
\midrule
DB & 43.28 & 54.36 & 94.20 & 48.77 & 181.87 & 50.89 \\
TB & 34.52 & 23.11 & 78.41 & 23.83 & 149.24 & 23.82 \\
HB & \textbf{34.01} & \textbf{21.29} & \textbf{76.67} & \textbf{21.08} & \textbf{144.48} & \textbf{19.87} \\
\bottomrule
\end{tabular}
}
\end{minipage}
\label{tab:HB}
\vspace{-4mm}
\end{table*}

\paragraph{Comparison of Balance Strategies.} To further validate the effectiveness of Hybrid Balance (HB), we conduct a comparison against Trajectory Balance (TB) and Detailed Balance (DB) under identical training settings on 100-node instances, evaluated on CVRP200, CVRP500, and CVRP1000. As shown in Table~\ref{tab:HB}, the HB module consistently outperforms both TB and DB across all instance sizes for both AGFN and GFACS. Notably, HB achieves up to a 15.07\% improvement over TB in AGFN and up to 16.58\% in GFACS. These results highlight the superior effectiveness of Hybrid Balance as a unifying optimization strategy.

\begin{table*}[t]
\centering
\vspace{-2mm}
\caption{Ablation Study on Depot-Guided Inference. Gap(\%) is computed with respect to the  LKH. DG represents depot greedy, DS represents depot sampling, CG represents customer greedy, CS represents customer sampling.}
\label{tab:Depot Guide-Inference}
\subcaption{AGFN}
\label{tab:Depot Guide-Inference_AGFN}
\resizebox{\textwidth}{!}{
\begin{tabular}{l|ccc|ccc|ccc}
\toprule
\multirow{2}{*}{\textbf{Method}} & \multicolumn{3}{c|}{\textbf{200}} & \multicolumn{3}{c|}{\textbf{500}} & \multicolumn{3}{c}{\textbf{1000}} \\
& \textbf{Obj. $\downarrow$} & \textbf{Time(s) $\downarrow$} & \textbf{Gap(\%) $\downarrow$} & \textbf{Obj. $\downarrow$} & \textbf{Time(s) $\downarrow$} & \textbf{Gap(\%) $\downarrow$} & \textbf{Obj. $\downarrow$} & \textbf{Time(s) $\downarrow$} & \textbf{Gap(\%) $\downarrow$} \\
\midrule
LKH & 28.04 & 59.81 & - & 63.32 & 233.72 & - & 120.53 & 433.90 & - \\
DG and CS & 32.78 & 0.16 & 16.90 & 76.42 & 0.41 & 20.69 & 146.14 & 0.65 & 21.25\\
DG and CG & 31.96 &0.16 & 13.98 & 71.35 & 0.41 & 12.68 &133.32 &0.65 & 10.61 \\
DS and CS & 31.88 & 0.16 & 13.69 & 74.49 & 0.41 & 17.64 & 144.74 & 0.65 & 20.09 \\
DS and CG & \textbf{30.83} & 0.16 & \textbf{9.95} & \textbf{69.93} & 0.41 & \textbf{10.44} & \textbf{131.78} & 0.65 & \textbf{9.34} \\
\bottomrule
\end{tabular}
}
\subcaption{GFACS}
\label{tab:Depot Guide-Inference_gfacs}
\resizebox{\textwidth}{!}{
\begin{tabular}{l|ccc|ccc|ccc}
\toprule
\multirow{2}{*}{\textbf{Method}} & \multicolumn{3}{c|}{\textbf{200}} & \multicolumn{3}{c|}{\textbf{500}} & \multicolumn{3}{c}{\textbf{1000}} \\
& \textbf{Obj. $\downarrow$} & \textbf{Time(s) $\downarrow$} & \textbf{Gap(\%) $\downarrow$} & \textbf{Obj. $\downarrow$} & \textbf{Time(s) $\downarrow$} & \textbf{Gap(\%) $\downarrow$} & \textbf{Obj. $\downarrow$} & \textbf{Time(s) $\downarrow$} & \textbf{Gap(\%) $\downarrow$} \\
\midrule
LKH & 28.04 & 59.81 & - & 63.32 & 233.72 & - & 120.53 & 433.90 & - \\
DG and CS & 34.87 & 4.67 & 24.36 & 75.99 & 12.78 & 20.01 & 148.53 & 26.33 & 23.23\\
DG and CG & 34.58 &4.67 & 23.32 & 74.12 & 12.78 & 17.06 &135.33 &26.33 & 12.28 \\
DS and CS & 33.38 & 4.67 & 19.04 & 75.79 & 12.78 & 19.69 & 143.46 & 26.33 & 19.02 \\
DS and CG & \textbf{32.66} & 4.67 & \textbf{16.48} & \textbf{71.89} & 12.78 & \textbf{13.53} & \textbf{133.32} & 26.33 & \textbf{10.61} \\
\bottomrule
\end{tabular}
}
\vspace{-2mm}
\end{table*}
\paragraph{Depot-Guided Inference Variants.} We assess four variants of the depot-guided inference strategy by applying either sampling or greedy decoding at the depot and customer nodes. Tests are conducted using both AGFN and GFACS on CVRP200, CVRP500, and CVRP1000. As shown in Table~\ref{tab:Depot Guide-Inference}, the combination of sampling at the depot and greedy decoding at customers yields the best performance. This setting consistently outperforms all other variants, including depot greedy + customer sampling, depot greedy + customer greedy, and depot sampling + customer sampling. These results validate the effectness of our depot-guided inference mechanism.

\begin{table*}[t]
\centering
\caption{Comparison of performance and runtime on TSP with 200, 500, and 1000 nodes. Gap(\%) is computed relative to LKH (10000).}
\resizebox{\textwidth}{!}{
\begin{tabular}{l|ccc|ccc|ccc}
\toprule
\multirow{2}{*}{\textbf{Method}} & \multicolumn{3}{c|}{\textbf{200}} & \multicolumn{3}{c|}{\textbf{500}} & \multicolumn{3}{c}{\textbf{1000}} \\
& \textbf{Obj. $\downarrow$} & \textbf{Time(s) $\downarrow$} & \textbf{Gap(\%) $\downarrow$} & \textbf{Obj. $\downarrow$} & \textbf{Time(s) $\downarrow$} & \textbf{Gap(\%) $\downarrow$} & \textbf{Obj. $\downarrow$} & \textbf{Time(s) $\downarrow$} & \textbf{Gap(\%) $\downarrow$} \\
\midrule
LKH & 10.62 & 38.80 & -- & 16.30 & 75.29 & -- & 22.68 & 149.36 & -- \\
ACO &45.72 &1.79 & 330.51 &149.62 &5.87 & 817.91 & 315.42 & 13.20 & 1290.74\\
\midrule
POMO & 10.97 & 0.12 & 3.30 & 20.85 & 0.39 & 27.91 & 33.94 & 0.59 & 49.65 \\
POMO (*8) & 10.90 & 0.20 & 2.64 & 20.44 & 0.55 & 25.40 & 32.60 & 3.42 & 43.74 \\
NeuOpt & 13.22 & 6.39 & 24.48 & 138.15 & 14.54 & 747.55 & 325.28 & 27.84 & 1334.22 \\
GANCO & 11.30 & 0.11 & 6.40 & 19.69 & 0.36 & 20.80 & 29.97 & 0.85 & 32.14 \\
\midrule
AGFN & 11.85 & 0.08 & 11.58 & 19.08 & 0.26 & 17.06 & 27.15 & 0.70 & 19.71 \\
Our-AGFN & \textbf{11.73} & 0.11 & \textbf{10.45} & \textbf{18.59} & 0.27 & \textbf{14.05} & \textbf{26.87} & 0.71 & \textbf{18.47} \\
GFACS & 13.04 & 1.64 & 22.79 & 24.41 & 9.42 &49.76 & 41.86 & 20.79 & 84.57 \\
Our-GFACS & \textbf{12.68} & 1.66 & \textbf{19.40} & \textbf{24.19} & 9.43 & \textbf{48.41} & \textbf{39.90} & 20.81 & \textbf{75.93} \\
GFACS (local search) & 10.78 & 6.67 & 1.51 & 17.10 & 27.76 & 4.91 & 24.45 & 58.42 & 7.80 \\
Our-GFACS (local search) & \textbf{10.78} & 6.68 & \textbf{1.50} & \textbf{17.05} & 27.78 & \textbf{4.60} & \textbf{24.42} & 58.42 & \textbf{7.67} \\
\bottomrule
\end{tabular}
}
\vspace{-6mm}
\label{tab:tsp_obj_time_gap}
\end{table*}

\subsection{Generalization to Other Vehicle Routing Problem} 
We further evaluate our framework on the Traveling Salesman Problem (TSP), a key VRP variant. Baselines include GFlowNet-based solvers (AGFN \cite{zhangadversarial}, GFACS \cite{kimant}), classical heuristics (LKH \cite{helsgaun2000effective}, ACO \cite{bell2004ant}), and learning-based models (POMO \cite{kwon2020pomo}, GANCO \cite{xin2022generative}, NeuOpt \cite{ma2024learning}). All models are trained on 100-node instances and evaluated on 200-, 500-, and 1,000-node settings. Since TSP lacks a depot node, depot-guided inference is not used. Table~\ref{tab:tsp_obj_time_gap} shows that HBG-AGFN consistently outperforms AGFN, reducing the gap by up to 17.64\%. HBG-GFACS also achieves notable improvements, with the gap on 200-node instances reduced from 22.79\% to 19.40\%. With local search, HBG-GFACS achieves further improvements, with the best gap reduction reaching 6.31\%. Compared to classical heuristics and learning-based methods such as ACO, POMO, NeuOpt, and GANCO, both HBG-AGFN and HBG-GFACS achieve competitive results on TSP tasks. These results confirm the generalizability and strong performance of HBG on TSP tasks.

\section{Conclusion}
In this paper, we introduced the Hybrid-Balance GFlowNet (HBG) framework to enhance the performance of GFlowNet-based solvers for vehicle routing problems. HBG unifies Trajectory Balance and Detailed Balance in a principled and adaptive manner to jointly optimize local and global objectives. We also proposed a depot-guided inference strategy aligned with the Hybrid-Balance principle, specifically tailored for depot-centric problems. Extensive experiments on both CVRP and TSP benchmarks demonstrate that HBG significantly improves the performance of two representative GFlowNet-based solvers, i.e., AGFN and GFACS, showcasing improved solution quality, scalability, and generalization. A current limitation of HBG is its reliance on existing GFlowNet-based models, as its performance depends in part on the underlying solver, which might be inferior to others. In future work, we plan to extend the Hybrid-Balance mechanism to a broader range of combinatorial optimization tasks and explore its integration with alternative generative policies and solvers.
\section*{Acknowledgments and Disclosure of Funding}
This research is supported by the National Research Foundation, Singapore under its AI Singapore Programme (AISG Award No: AISG3-RP-2022-031).
\bibliographystyle{plain} 
\bibliography{references}

\newpage
\appendix
\section{Methodology Details}
\subsection{Graph Neural Network}
\label{sec:GNN}
Our GNN architecture follows the same design as those used in AGFN and GFACS to ensure a fair comparison, which uses a custom GNN architecture designed specifically for the VRP task, and employs a custom message-passing GNN that jointly updates node and edge representations over multiple layers. 
At each layer $l$, node embedding $h_i^l$ of the $i$-th node and edge embedding $e_{ij}^l$ between the $i$-th and $j$-th nodes 
are updated via the following formulations:
\begin{align}
h_i^{l+1} &= h_i^l + \text{ACT}\Big( 
\text{BN}\big( 
W_1^l h_i^l + A(\sigma(e_{ij}^l) \odot W_2^l h_j^l)
\big) \Big), \\
e_{ij}^{l+1} &= e_{ij}^l + \text{ACT}\Big( 
\text{BN}\big( 
W_3^l e_{ij}^l + W_4^l h_i^l + W_5^l h_j^l
\big) \Big),
\end{align}
where $W_1^l, W_2^l, W_3^l, W_4^l, W_5^l$ are learnable parameters, 
$A(\cdot)$ denotes the aggregation function (mean pooling in our case), 
$\sigma(\cdot)$ is the sigmoid activation that modulates attention over neighbors, 
and $\text{ACT}$ denotes the SiLU activation. 
Batch normalization (BN) is applied at each step for stability. 
The number of layers is set to 12 for HBG-GFACS and 16 for HBG-AGFN, 
with hidden dimensions of 32 and 64, respectively.
\subsection{Proof Process of Hybrid-Balance}
\label{sec:Proof Process of Hybrid-Balance}
To solve Eq. \ref{eq. Trajectory Count Recurrence}, We begin by establishing the boundary conditions. When either $a = 0$ or $j = 0$, the recurrence simplifies accordingly. For instance, when $a = 0$, we obtain:
\begin{equation}
B(\mathcal{A}_0, \mathcal{J}_j) = j \cdot B(\mathcal{A}_0, \mathcal{J}_{j-1}).
\label{eq:j=0expression}
\end{equation}
When $a = 0$ and $j = 0$, there are no sub-trajectories in $\tau_i$, and $\tau_i$ contains only the depot node. Therefore, the base case becomes $B(\mathcal{A}_0, \mathcal{J}_0) = 1$, and it is equivalent to:
\begin{equation}
B(\mathcal{A}_0, \mathcal{J}_0) = 1 = \frac{(0 + 0)!}{0! \cdot 0!}, \quad \text{for } a = j = 0.
\label{eq:a,j=0}
\end{equation}
Using Eq.~\ref{eq:j=0expression} recursively and applying the base case $B(\mathcal{A}_0, \mathcal{J}_0) = 1$, we derive:
\begin{equation}
B(\mathcal{A}_0, \mathcal{J}_j) = j!.
\label{eq: a=0}
\end{equation}
Similarly, when $j = 0$, we can deduce that:
\begin{equation}
B(\mathcal{A}_a, \mathcal{J}_0) = 2^a \cdot a!.
\label{eq: j=0}
\end{equation}

We now normalize Eq.~\ref{eq. Trajectory Count Recurrence} by dividing both sides by $2^a \cdot a! \cdot j!$, resulting in:
\begin{equation}
\frac{B(\mathcal{A}_a, \mathcal{J}_j)}{2^a \cdot a! \cdot j!} = 
\frac{B(\mathcal{A}_{a-1}, \mathcal{J}_j)}{2^{a-1} \cdot (a - 1)! \cdot j!} + 
\frac{B(\mathcal{A}_a, \mathcal{J}_{j-1})}{2^a \cdot a! \cdot (j - 1)!}.
\label{eq:b_divide}
\end{equation}
Based on this, we define a normalized function:
\begin{equation}
c(\mathcal{A}_a, \mathcal{J}_j) \triangleq \frac{B(\mathcal{A}_a, \mathcal{J}_j)}{2^a \cdot a! \cdot j!}.
\label{eq:c_definition}
\end{equation}
Substituting Eq.~\ref{eq:c_definition} into Eq.~\ref{eq:b_divide}, we obtain the recurrence:
\begin{equation}
c(\mathcal{A}_a, \mathcal{J}_j) = c(\mathcal{A}_{a-1}, \mathcal{J}_j) + c(\mathcal{A}_a, \mathcal{J}_{j-1}).
\label{eq: c_recurrence}
\end{equation}

We analyze the recurrence in Eq.~\ref{eq: c_recurrence} under the boundary conditions, and derive results from Eqs.~\ref{eq: a=0}, \ref{eq: j=0}, and \ref{eq:c_definition}:
\[
c(\mathcal{A}_0, \mathcal{J}_j) = 1, \quad c(\mathcal{A}_a, \mathcal{J}_0) = 1, \quad \text{for all } a, j \geq 0.
\]
These boundary conditions are equivalent to:
\begin{equation}
c(\mathcal{A}_0, \mathcal{J}_j) = 1 = \frac{(0 + j)!}{0! \cdot j!}, \quad \text{for } a = 0, j > 0,
\label{eq:ini_1}
\end{equation}
\begin{equation}
c(\mathcal{A}_a, \mathcal{J}_0) = 1 = \frac{(a + 0)!}{a! \cdot 0!}, \quad \text{for } a > 0, j = 0.
\label{eq:ini_2}
\end{equation}

Moreover, motivated by Pascal’s rule \cite{graham1994concrete}, which states:
\[
\binom{n}{k} = \binom{n - 1}{k - 1} + \binom{n - 1}{k}, \quad 
\text{for all integers } n \geq 1 \text{ and } 1 \leq k \leq n,
\]
and by setting \( n = a + j \), \( k = a \), we obtain:
\[
\binom{a + j}{a} = \binom{a + j - 1}{a - 1} + \binom{a + j - 1}{a}, \quad \text{for } a, j \geq 1,
\]
which is equivalent to:
\[
\binom{a + j}{a} = \binom{(a - 1) + j}{a - 1} + \binom{a + (j - 1)}{a}.
\]
This is similar with the function of $c(\mathcal{A}_a, \mathcal{J}_j)$ in Eq.~\ref{eq: c_recurrence}. Therefore, the closed-form expression of $c(\mathcal{A}_a, \mathcal{J}_j)$ is:
\begin{equation}
c(\mathcal{A}_a, \mathcal{J}_j) = \binom{a + j}{a} = \frac{(a + j)!}{a! \cdot j!}, \quad \text{for } a, j \geq 1.
\label{eq: c_expression 2}
\end{equation}
By extending Eqs.~\ref{eq:a,j=0}, \ref{eq:ini_1}, \ref{eq:ini_2}, and \ref{eq: c_expression 2}, the generalized closed-form solution for all $a, j \geq 0$ is given as:
\begin{equation}
c(\mathcal{A}_a, \mathcal{J}_j) = \binom{a + j}{a} = \frac{(a + j)!}{a! \cdot j!}.
\label{eq: c_expression}
\end{equation}

Finally, substituting Eq.~\ref{eq: c_expression} into Eq.~\ref{eq:c_definition}, we obtain the closed-form expression for the total number of distinct sub-trajectory orderings resulting in the same complete trajectory:
\begin{equation}
B(\mathcal{A}_a, \mathcal{J}_j) = (a + j)! \cdot 2^a, \quad \text{for } a, j \geq 0.
\end{equation}
\section{More Experiments}
\subsection{Test on Real world dataset}
\begin{table*}[t]
\centering
\caption{Test on the CVRPLIB-XXL benchmark. Gap(\%) is computed with respect to the optimal solution.}
\label{tab:CVRPLib-xxl}
\resizebox{\textwidth}{!}{
\begin{tabular}{l|cccccccc}
\toprule
\textbf{Gap(\%) $\downarrow$} & \textbf{L1 (3k)} & \textbf{L2 (4k)} & \textbf{A1 (6k)} & \textbf{A2 (7k)} & \textbf{G1 (10k)} & \textbf{G2 (11k)} & \textbf{B1 (15k)} & \textbf{B2 (16k)} \\
\midrule
AGFN & 1145.18 & 27.87 & 29.62 & 24.59 & 142.62 & 27.24 & \textbf{20.65} & 120.69 \\
\textbf{HBG-AGFN} & \textbf{97.15} & \textbf{25.09} & \textbf{25.36} & \textbf{20.96} & \textbf{113.89} & \textbf{20.47} & 109.80 & \textbf{46.97} \\
GFACS & 119.24 & 1788.57 & 112.08 & 236.79 & 281.34 & 2207.71 & 352.69 & 2537.29 \\
\textbf{HBG-GFACS} & \textbf{52.25} & \textbf{33.96} & \textbf{21.58} & \textbf{39.19} & \textbf{78.88} & \textbf{31.60} & \textbf{179.70} & \textbf{37.43} \\
\bottomrule
\end{tabular}
}
\end{table*}

\label{sec:Test on Real world dataset}
To evaluate our model's performance on real-world data, we conduct experiments on the CVRPLIB-XXL ~\cite{arnold2019efficiently}, which is designed to test model's performance on large-scale real-world instances. As shown in Table~\ref{tab:CVRPLib-xxl}, HBG-AGFN achieves up to a 91.51\% reduction in gap compared to AGFN, while HBG-GFACS achieves up to 98.52\% improvement over GFACS. These results demonstrate that our framework significantly enhances the performance of GFlowNet-based solvers on real-world datasets.

\begin{table*}[h]
\centering
\caption{Comparison of Hyperparameter Settings on CVRP}
\label{tab:combined_cvrp}
\begin{subtable}[t]{0.32\textwidth}
\centering
\caption{Sparsity Parameter}
\label{tab:sparsity}
\resizebox{\textwidth}{!}{
\begin{tabular}{lccc}
\toprule
\textbf{CVRP (Gap\%)} & 200 & 500 & 1000 \\
\midrule
HBG\_GFACS\_2 & 19.26 & 16.27 & 12.69 \\
HBG\_GFACS\_5(origin) & \textbf{16.48} & 13.59 & 10.61 \\
HBG\_GFACS\_8 & 17.90 & 13.83 & 11.02 \\
HBG\_GFACS\_10 & 20.43 & 17.88 & 14.9 \\
HBG\_GFACS(local)\_5(origin) & 1.96 & \textbf{2.81} & 2.75 \\
HBG\_GFACS(local)\_10 & 2.07 & 3.43 & 4.57 \\
HBG\_AGFN\_2 & 12.62 & 13.48 & 17.32 \\
HBG\_AGFN\_5(origin) & 9.50 & \textbf{10.44} & 9.34 \\
HBG\_AGFN\_8 & 11.95 & 13.57 & 12.69 \\
HBG\_AGFN\_10 & 9.10 & 14.56 & 14.04 \\
\bottomrule
\end{tabular}}
\end{subtable}
\hfill
\begin{subtable}[t]{0.32\textwidth}
\centering
\caption{Learning Rate}
\label{tab:learningrate}
\resizebox{\textwidth}{!}{
\begin{tabular}{lccc}
\toprule
\textbf{CVRP (Gap\%)} & 200 & 500 & 1000 \\
\midrule
HBG\_GFACS\_5$\times10^{-3}$ & 17.15 & 14.48 & 11.77 \\
HBG\_GFACS\_1$\times10^{-3}$ & 17.43 & 14.57 & 11.50 \\
HBG\_GFACS\_5$\times10^{-4}$(origin) & \textbf{16.48} & \textbf{13.59} & \textbf{10.61} \\
HBG\_GFACS\_1$\times10^{-4}$ & 18.22 & 14.19 & 11.30 \\
HBG\_GFACS(local)\_5$\times10^{-3}$ & 1.71 & 2.77 & 4.33 \\
HBG\_GFACS(local)\_1$\times10^{-3}$ & \textbf{1.70} & \textbf{2.65} & \textbf{2.51} \\
HBG\_GFACS(local)\_5$\times10^{-4}$(origin) & 1.93 & 4.42 & 5.75 \\
HBG\_GFACS(local)\_1$\times10^{-4}$ & 4.53 & 6.49 & 6.50 \\
HBG\_AGFN\_5$\times10^{-3}$ & 11.41 & 14.62 & 11.00 \\
HBG\_AGFN\_1$\times10^{-3}$ & 10.14 & 12.77 & 10.62 \\
HBG\_AGFN\_5$\times10^{-4}$(origin) & \textbf{9.95} & \textbf{10.44} & \textbf{9.34} \\
HBG\_AGFN\_1$\times10^{-4}$ & 9.37 & 10.27 & 9.59 \\
\bottomrule
\end{tabular}}
\end{subtable}
\hfill
\begin{subtable}[t]{0.32\textwidth}
\centering
\caption{Optimizer Type}
\label{tab:optimizer}
\resizebox{\textwidth}{!}{
\begin{tabular}{lccc}
\toprule
\textbf{CVRP (Gap\%)} & 200 & 500 & 1000 \\
\midrule
HBG\_GFACS\_SGD & 17.80 & 15.19 & 12.45 \\
HBG\_GFACS\_Adam & 17.33 & 14.28 & 11.11 \\
HBG\_GFACS\_AdamW(origin) & \textbf{16.48} & \textbf{13.59} & \textbf{10.61} \\
HBG\_GFACS(local)\_SGD & 4.23 & 4.26 & 4.55 \\
HBG\_GFACS(local)\_Adam & \textbf{2.03} & 2.70 & 2.73 \\
HBG\_GFACS(local)\_AdamW(origin) & 2.30 & 3.43 & 4.57 \\
HBG\_AGFN\_SGD & 14.66 & 13.00 & 12.47 \\
HBG\_AGFN\_Adam & 10.16 & 11.88 & 11.77 \\
HBG\_AGFN\_AdamW(origin) & \textbf{9.95} & \textbf{10.44} & \textbf{9.34} \\
\bottomrule
\end{tabular}}
\end{subtable}
\end{table*}

\subsection{Hyperparameter Sensitivity Analysis on CVRP}
The introduction of a new loss component ($L_{DB}$) could potentially alter the optimization landscape, and that the original hyperparameter settings used in AGFN and GFACS may not be optimal for the proposed HBG variants. We further conduct additional experiments in which we re-tuned key hyperparameters for the HBG models, including the sparsity parameter, learning rate, and optimizer settings. The updated results (Table~\ref{tab:sparsity}--\ref{tab:optimizer}) will be incorporated into the Appendix of the revised version. HBG-GFACS\_2 refers to the HBG-GFACS model evaluated with a sparsity parameter $k=|V|/2$; the interpretation is analogous for the other entries. 
The term \textit{origin} denotes the original hyperparameter configuration used in our main experiments. 
HBG-GFACS\_5$\times10^{-3}$ refers to the HBG-GFACS model evaluated with a learning rate of $5\times10^{-3}$; the interpretation is analogous for the other entries. 
HBG-GFACS\_SGD refers to the HBG-GFACS model evaluated using the SGD optimizer; the interpretation is analogous for the other entries.

As the sparsity parameter results shown in Table~\ref{tab:sparsity}, the optimal sparsity setting yields the best performance for both HBG-GFACS and HBG-AGFN. Regarding the learning rate comparisons in Table~\ref{tab:learningrate}, for HBG-GFACS, the original value of $5\times 10^{-4}$ performs best without local search, while a value of $1\times 10^{-3}$ achieves the best results when local search is enabled. For HBG-AGFN, a learning rate of $1\times 10^{-4}$ shows superior performance on the 200- and 500-node instances, whereas the original value performs best on the 1000-node instances. As for the optimizer comparison in Table~\ref{tab:optimizer}, the original AdamW setting provides the best performance in most cases. An exception is observed on the 500- and 1000-node instance with local search, where Adam slightly outperforms AdamW for HBG-GFACS.

\subsection{Weight $\lambda$ of DB and TB}
\begin{table*}[ht]
\centering
\caption{Weights Analysis of HBG-AGFN on CVRP}
\label{tab:weight_cvrp}
\resizebox{\textwidth}{!}{
\begin{tabular}{lcccccccccc}
\toprule
\textbf{Size (Gap\%)} & 0.5$\to$0 & 1$\to$0 & 2$\to$0 & 0.5$\to$0.5 & 1$\to$0.5 & 2$\to$0.5 & 0.5$\to$1 & 1$\to$1 (origin) & 2$\to$2 \\
\midrule
200 & 11.09&10.80 & 10.94 & 11.41 & 11.98 & 11.20 & 9.98 & \textbf{9.95} & 10.44 \\
500 & 11.27&12.04 & 11.18 &12.15& 11.66 & 11.80 & 11.63 & \textbf{10.44} & 11.98 &  \\
1000 & 10.89 & 10.61 & 9.89 & 11.03 & 10.84 & 10.98 &10.87& \textbf{9.34} & 10.08 \\
\bottomrule
\end{tabular}}
\end{table*}

\begin{table*}[ht]
\centering
\caption{Weights Analysis of HBG-AGFN on TSP}
\label{tab:weight_tsp_agfn}
\resizebox{\textwidth}{!}{
\begin{tabular}{lcccccccccc}
\toprule
\textbf{Size (Gap\%)} & 0.5$\to$0 & 1$\to$0 & 2$\to$0 & 0.5$\to$0.5 & 1$\to$0.5 & 2$\to$0.5 & 0.5$\to$1 & 1$\to$1 (origin) & 2$\to$2 \\
\midrule
200 & 10.26 & 10.42 & 10.49 & 11.48 & 11.26 & 10.66 & \textbf{10.08} & 10.45&10.36\\
500 & 15.15 & 15.40 & 16.13 & 16.87 & 16.09 & 15.79&14.91& \textbf{14.05} & 15.64 \\
1000 & 18.80 & 18.55 & 19.00 &19.26& 21.87 & 22.00 & 18.52 & \textbf{18.47} & 18.81 \\
\bottomrule
\end{tabular}}
\end{table*}

\begin{table*}[ht]
\centering
\caption{Weights Analysis of HBG-GFACS on CVRP}
\label{tab:weight_cvrp_gfacs}
\resizebox{\textwidth}{!}{
\begin{tabular}{lcccccccccc}
\toprule
\textbf{Size / Gap(\%) / $\lambda$} 
& 0.5$\to$0 & 1$\to$0 & 2$\to$0 
& 0$\to$0.5 & 0$\to$1 & 0$\to$2 
& 0.5$\to$0.5 & 1$\to$1 (origin) & 2$\to$2 \\
\midrule
200 & \textbf{14.05} & 19.29 & 17.37 & 17.72 & 18.47 & 19.08 & 16.42 & 16.48 & 18.87 \\
500 & \textbf{11.78} & 17.66 & 15.38 & 15.28 & 16.35 & 15.71 & 13.57 & 13.53 & 15.49 \\
1000 & 11.29 & 15.80 & 12.71 & 12.88 & 13.47 & 12.80 & 10.91 & \textbf{10.61} & 12.80 \\
200 (local search) & 2.39 & \textbf{1.78} & 2.03 & 2.07 & 2.14 & 1.96 & 1.96 & 1.96 & 1.78 \\
500 (local search) & 3.02 & \textbf{2.51} & 2.91 & 3.41 & 3.44 & 2.99 & 3.17 & 2.81 & 2.62 \\
1000 (local search) & 2.82 & \textbf{2.27} & 2.82 & 3.31 & 3.53 & 3.22 & 2.75 & 2.75 & 2.40 \\
\bottomrule
\end{tabular}}
\end{table*}

\begin{table*}[ht]
\centering
\caption{Weights Analysis of HBG-GFACS on TSP}
\label{tab:weight_tsp_gfacs}
\resizebox{\textwidth}{!}{
\begin{tabular}{lcccccccccc}
\toprule
\textbf{Size / Gap(\%) / $\lambda$} 
& 0.5$\to$0 & 1$\to$0 & 2$\to$0 
& 0$\to$0.5 & 0$\to$1 & 0$\to$2 
& 0.5$\to$0.5 & 1$\to$1 (origin) & 2$\to$2 \\
\midrule
200 & \textbf{19.13} & 20.61 & 23.06 & 22.21 & 20.56 & 24.02 & 21.64 & 19.40 & 22.82 \\
500 & 49.51 & 49.61 & 50.80 & 49.48 & 48.47 & 56.13 & 49.62 & \textbf{48.41} & 50.37 \\
1000 & \textbf{72.30} & 80.19 & 90.50 & 79.37 & 80.39 & 96.36 & 77.45 & 75.93 & 87.23 \\
200 (local search) & 1.57 & \textbf{1.49} & 1.59 & 1.59 & 1.62 & 1.82 & 1.54 & 1.50 & 1.62 \\
500 (local search) & 4.81 & \textbf{4.65} & 5.02 & 4.88 & 4.73 & 5.18 & 4.73 & 4.68 & 5.12 \\
1000 (local search) & 7.73 & 7.68 & 7.86 & 7.83 & 8.02 & 8.20 & 7.70 & \textbf{7.67} & 7.99 \\
\bottomrule
\end{tabular}}
\end{table*}

We have conducted additional experiments on the weighted combination (i.e., $\mathcal{L}_{\mathrm{HB}} = \mathcal{L}_{\mathrm{TB}} + \lambda \mathcal{L}_{\mathrm{DB}}$) to assess the sensitivity of our method to different values of $\lambda$, which are gathered in Tables~\ref{tab:weight_cvrp}–\ref{tab:weight_tsp_gfacs}. 
The results include both adaptive weights (i.e., $\lambda$: 0.5$\to$0, 1$\to$0, 2$\to$0 and so on) and fixed weights (i.e., 0.5$\to$0.5, 1$\to$1, 2$\to$2), where 0.5$\to$0 denotes changing the TB:DB loss weight ratio from 1:0.5 to 1:0 during training, with TB weight fixed at 1. The interpretation for other entries follows analogously. The term \textit{origin} denotes the original hyperparameter setting used in our main experiments.
We find that, for HBG-AGFN, a fixed 1:1 weight between TB and DB consistently yields the best performance. 
Similarly, for HBG-GFACS, this ratio offers the most favorable trade-off between performance and stability with and without local search.

\subsection{Statistical Significance of Experiment}
\begin{figure}[htbp]
  \centering
  \begin{subfigure}{0.48\textwidth}
    \includegraphics[width=\linewidth]{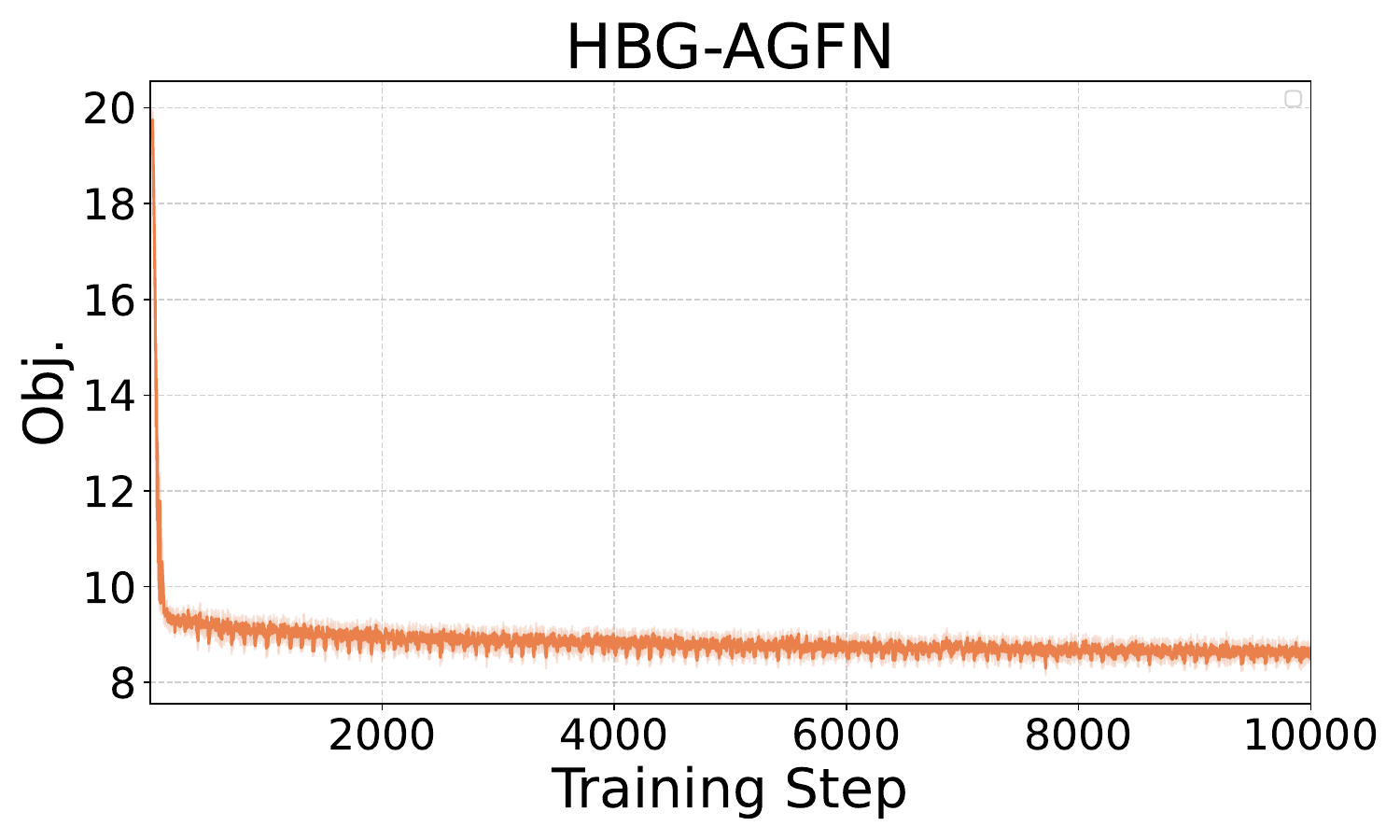}
  \end{subfigure}
  \hfill
  \begin{subfigure}{0.48\textwidth}
    \includegraphics[width=\linewidth]{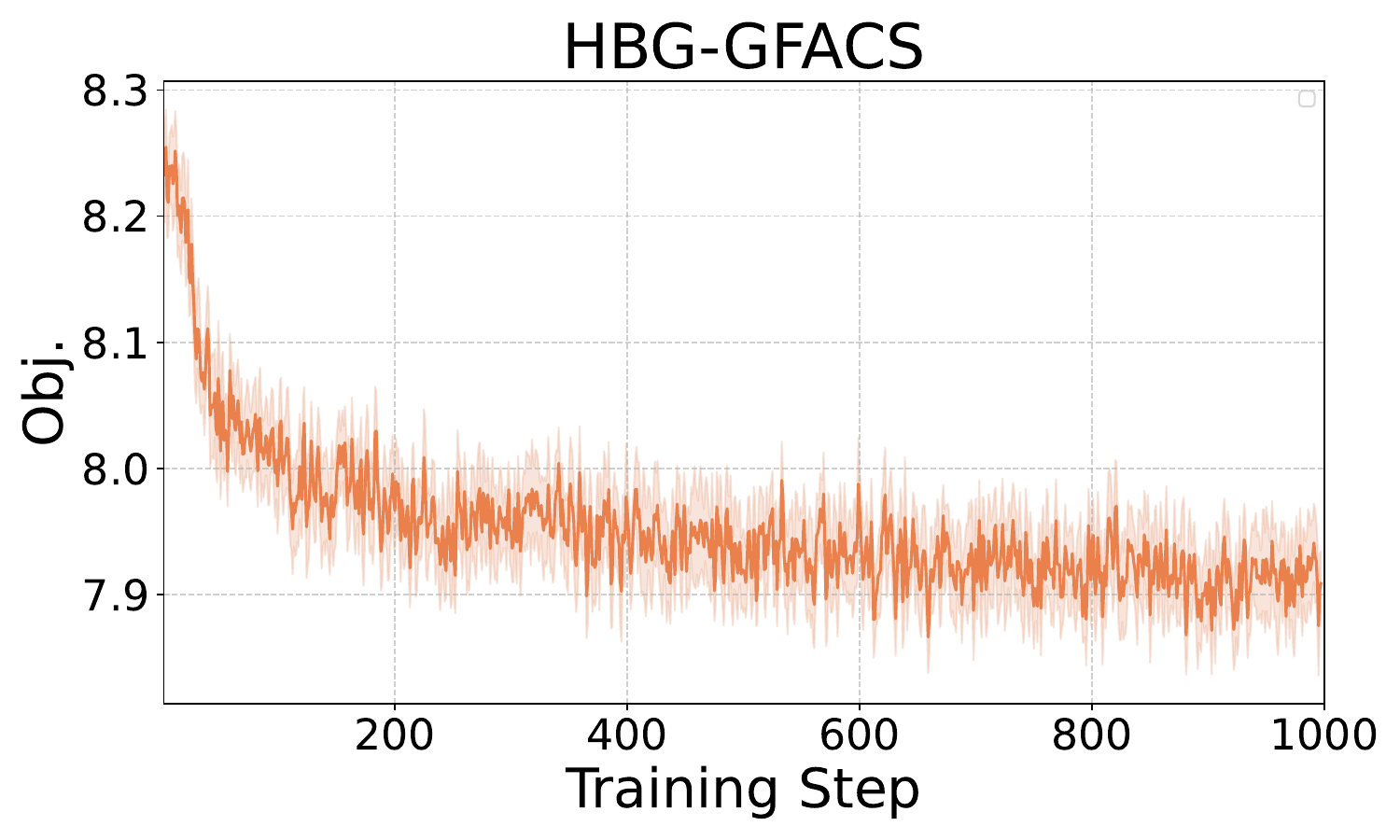}
  \end{subfigure}
  \caption{Training Process of HBG-AGFN and HBG-GFACS.}
  \label{fig:statistical_significance}
\end{figure}

\label{sec:Statistical Significance of Experiment}
As shown in Fig. \ref{fig:statistical_significance}, both HBG-AGFN and HBG-GFACS exhibit a clear and steady decline in objective values throughout the training process, indicating stable convergence. The narrow shaded areas—representing standard deviation across five random seeds—suggest low variance among runs. These results collectively highlight the effectiveness of the training process and the statistical reliability of the observed performance gains.

\end{document}